# Stabilization of Bipedal Robot Motion based on Total Momentum

Erfan Ghorbani[1], Venus Pasandi[1], Mehdi Keshmiri[1], and Mostafa Ghobadi[2]

[1] *Isfahan University of Technology, Isfahan, Iran*
[2] *State University of New York at Buffalo, Buffalo, US*

**Abstract** Bipedal robots adapt to the environment of the modern society due to the similarity of movement to humans, and therefore they are a good partner for humans. However, maintaining the stability of these robots during walking/running motion is a challenging issue that, despite the development of new technologies and the advancement of knowledge, does not yet have a satisfactory solution. In most of the proposed methods by researchers, to maintain the stability of walking bipedal robots, it has been tried to ensure the momentary stability of motion by limiting the motion to multiple constraints. Although these methods have good performance in sustaining stability, they leave the robot away from the natural movement of humans, with low efficiency and high energy consumption. Hence, many researchers have turned to the walking techniques that follow a certain motion limit cycle, in which we can consider the overall stability rather than momentary. In this paper, a method is proposed to maintain the stability of the limit cycle against disturbance. For this purpose, the dynamical model of the biped robot is extracted in the space of total momentum variables and, according to the desired step length and speed, the motion limit cycle is designed. Subsequently, a motion stabilizer is proposed based on the idea of length shift, which is a natural human strategy for sustaining the balance in case of impact. The simulations show that this technique has a good performance in maintaining the stability of motion and has similar responses to human response.

ISME۲۰۱۹-XXXX

# پایدارسازی حرکت ربات دوپا براساس اندازه حرکت کلی

عرفان قربانی[۱]، ونوس پسندی[۲]، مهدی کشمیری[۳]، مصطفی قبادی[۴]

[۱]دانشجوی کارشناسی‌ارشد، دانشگاه صنعتی اصفهان، erfan.ghorbani@me.iut.ac.ir
[۲]دانشجوی دکتری، دانشگاه صنعتی اصفهان، venus.pasandi@me.iut.ac.ir
[۳]استاد، دانشگاه صنعتی اصفهان، mehdik@cc.iut.ac.ir
[۴]دانشجوی دکتری، دانشگاه ایالتی نیویورک در بوفالو، mostafag@buffalo.edu

## چکیده

ربات‌های دوپا به دلیل شباهت حرکتی به انسان، با محیط جامعه مدرن سازگاری داشته و از این رو شریک مناسبی برای انسان محسوب می‌شوند. اگرچه، حفظ پایداری و کنترل راهرفتن این ربات‌ها مسئله‌ای چالش برانگیز است که علی‌رغم توسعه تکنولوژی و پیشرفت دانش هنوز راه حل مناسبی برای آن ارائه نشده است. در اکثر روش‌های پیشنهاد شده توسط محققین برای حفظ پایداری راهروندهای دوپا، سعی شده است تا با محدود کردن حرکت به قیدهای متعدد، پایداری حرکت در هر لحظه تضمین شود. اگرچه این روش‌ها عملکرد خوبی در حفظ پایداری دارند، اما راهرفتن ربات را از حرکت طبیعی انسان دور کرده و کارایی پایین و مصرف انرژی بالا را همراه دارند. از این رو، محققین بسیاری به تکنیک راه رفتن چرخه حدی[۱] روی آورده‌اند که پایداری کلی و نه لحظه‌ای حرکت را مد نظر قرار می‌دهد. در این مقاله روشی برای حفظ پایداری چرخه حدی در مقابل اغتشاش ارائه شده است. بدین منظور مدل دینامیکی ربات دوپا در فضای متغیرهای اندازه حرکت کلی استخراج شده و با توجه به طول و سرعت گام مطلوب، چرخه حرکتی طراحی می‌شود. پس از آن، براساس ایده تغییر طول گام، که از راهبردهای طبیعی انسان برای حفظ پایداری حرکت در مواجهه با ضربه است، یک پایدارساز حرکت پیشنهاد شده است. شبیه‌سازی‌های انجام شده نشان می‌دهد که این تکنیک کارایی مناسبی در حفظ پایداری حرکت داشته و واکنش‌هایی مشابه با واکنش انسان را درپی دارد.

## واژه های کلیدی

پایداری ربات دوپا، راهرفتن چرخه حدی، مقابله با اغتشاش، راهبرد تغییر طول گام

## مقدمه

یکی از پیچیده‌ترین و بلندمدت‌ترین برنامه‌های توسعه ربات‌ها دستیابی به ربات انسان‌نمای دارای همه توانایی‌های مورد نیاز برای همکاری و کمک به انسان در زندگی روزمره و محیط اطراف او است. اولین و اساسی‌ترین گام در تحقق این هدف، فراهم کردن توانایی راه رفتن این ربات‌هاست. راهرفتن یک ربات دوپا را می‌توان به صورت "حرکت در طول با سرعت متوسط به وسیله بلند کردن و پایین گذاشتن هر پا به نوبت به طوریکه یک پا در تماس با زمین است درحالیکه پای دیگر برداشته می‌شود" تعریف نمود [۱]. این ماهیت ترکیبی[۲] حرکت به علاوه دینامیک غیرخطی و همچنین درجات آزادی زیاد، راهرفتن و حفظ پایداری ربات دوپا را به چالشی اساسی تبدیل کرده است. ابتدایی‌ترین راهکار مطرح شده برای حل این چالش، راهرفتن با پایداری استاتیکی است، بدین معنا که تصویر عمودی مرکز جرم همواره در ناحیه تکیه‌گاهی بماند. این نوع پایداری، بسیار محدود کننده بوده و به راهرفتنی آرام با طول گام‌های کوتاه منجر می‌شود. بعد از آن، ووکبراداویچ با مطرح کردن معیار نقطه گشتاور صفر (ZMP)[۳] معیاری برای پایداری دینامیکی ارائه کرد [۲]. این معیار با مقید کردن کف‌پای تکیه‌گاهی به تماس دائمی با زمین از پایداری و نیفتادن ربات اطمینان حاصل می‌کند. با وجود اینکه معیار ZMP امکان حرکت با سرعت بیشتر و طول گام‌های بلندتری نسبت به راهرفتن استاتیکی را فراهم می‌کند، اما همچنان با مصرف انرژی بالاتر و چابکی پایین‌تری نسبت به راهرفتن انسان همراه است.

اولین‌بار هرموزلو در بررسی اثر ضربه کف‌پا در پایداری ربات دوپا، راهرفتن را به عنوان یک چرخه حدی تحلیل کرد [۳]. با این وجود پژوهش‌های مک‌گیر در ارتباط با راه رونده‌های غیرفعال[۴] (بدون عملگر) نقطه عطفی در زمینه راهرفتن چرخه حدی بود [۴]. هوبلن راهرفتن چرخه حدی را به صورت رسمی بدین صورت تعریف کرد که "راهرفتن چرخه حدی، دنباله‌ای از طول گام‌هاست که به طور کلی پایدار است اما به صورت محلی در هر لحظه از زمان پایدار نیست" [۵]. این الگوی راهرفتن به علت فراهم کردن امکان حرکت با سرعت‌های بالا، طول گام‌های بلند و مصرف انرژی پایین توجه بسیاری از محققین را جلب کرده است. هوبلن همچنین با توجه به اثر مثبت ضربه کف‌پای متحرک در پایدارسازی چرخه حدی در مقابل اغتشاشات، راهبرد پس‌کشی پای متحرک[۵] در انتهای گام را برای افزایش اثر آن مطرح کرد [۶]. بونسول با پیدا کردن چرخه‌های حدی و نواحی جذبشان برای مدل ساده پاندول معکوس، بر اساس اشتراک

---

[۲] Hybrid
[۳] Zero moment point
[۴] Passive
[۵] Swing leg retraction

[۱] Limit cycle walking

۱

نواحی جذب روشی برای گذار بین چرخه‌ها به عنوان مدلی از دویدن ربات دوپا ارائه کرد [7]. او در کار دیگری برای غلبـه بـر اغتشـاش در ربات دوپای کم‌عملگر راهبرد پیش‌کشی پای متحرک[6] را مطرح کرد و نشان داد استراتژی مناسب بین دو راهبرد پس‌کشی یا پیش‌کشی پای متحرک به عواملی چون شیب سطح، تغییر سرعت در اثر اغتشاش و... بستگی دارد [8]. سالمن برای مدل پنج لینکی ربات دوپای کم‌عملگر کنترل‌کننده‌های خطی پیشنهاد داد که با تنظیم پارامترهای‌کنترلی به وسیله الگوریتم جستجوی تکاملی، توانایی تطبیق با تغییر شیب سطح و مقابله با اغتشاشات ناشی از تغییر ارتفاع سطح را دارد [9].

در ادامه این مقاله ابتـدا مـدل دینـامیکی ربـات دوپـا در فضـای متغیرهای اندازه حرکت کلی، شامل اندازه حرکت خطی و زاویـه‌ای، استخراج و ساده‌سازی شده است. سپس برای دینامیـک سـاده شـده، چرخه حرکتی متناسب با طول و سرعت گام مطلوب بدسـت آمـده و پایداری آن بررسی شده است. بعد از آن یک پایدارساز تغییر طول گام برای حفظ پایداری حرکت به هنگام مواجهه بـا اغتشـاش طراحـی و ماتریس ضرایب کنترلی بهینه‌سازی شده است. در انتهـا عملکـرد دو پایدارساز پیشنهاد شده با شبیه‌سازی بررسی شده است.

## مدل دینامیکی حرکت دوپا

راه‌رفتن ربات دوپا متشکل از دو فاز تک تکیه‌گاهی و دو تکیـه‌گاهی است. در فاز تک‌تکیه‌گاهی، یک پا (پای تکیه‌گاهی) در تماس دائمی با زمین و پای دیگر (پای متحرک) در هوا متحرک اسـت درحالیکـه در فاز دو تکیه‌گاهی، هر دو پا در تماس با زمین است. در انتهای هـر فـاز تک تکیه‌گاهی، فـاز دو تکیه‌گاهی رخ می‌دهـد و بـالعکس. از ایـن‌رو معادلات حرکت ربات دوپا ترکیبی از معادلات این دو فاز خواهد بود.

### فاز تک تکیه‌گاهی

برای استخراج معادلات در فاز تک تکیه‌گاهی، ربات دوپـا بـه صـورت یک سیستم کلی در تماس دائمی با زمین (شـکل 1) در نظـر گرفتـه می‌شود که دینامیک داخلی آن (تغییر زوایـای مفاصـل) مـورد توجـه نمی‌باشد. براساس قوانین نیوتون باتوجه بـه انـدرکنش بـین ربـات و زمین، معادلات حرکت فاز تک تکیه‌گاهی به صورت زیر است.

$$
\begin{cases}
f_x = m\ddot{x}_{COM} \\
f_y - mg = m\ddot{y}_{COM} \\
f_x y_{COM} - f_y x = \dot{H}
\end{cases}
\quad (1)
$$

تمامی متغیرها در فهرست علائم در انتهای مقاله معرفی شده است.
از طرفی باتوجه به حرکت طبیعی انسان فرض می‌شـود در طـی یک گام:

- ارتفاع مرکز جرم ثابت می‌ماند ($\dot{y}_{COM} = 0$).
- اندازه‌حرکت زاویه‌ای کل حول مرکز جـرم تغییـر نمی‌کنـد ($\dot{H} = 0$).
- شتاب تغییر موقعیت مرکز فشار صفر است ($\ddot{x} = \ddot{x}_{COM}$).

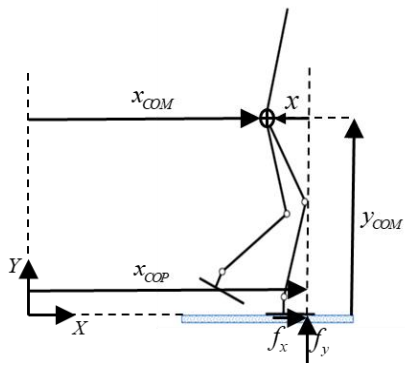

شکل 1: مدل ربات دوپا

براین اساس معادله دیفرانسیل خطی زیر به عنوان مـدل سـاده شـده حرکت ربات دوپا بدست آمده و مرتبه سیستم به دو تقلیل می‌یابد.

$$\ddot{x} = \frac{g}{h} x \quad (2)$$

در ادامه این مقاله همواره از مدل ساده شده (2) استفاده شده است. معادله (2) همان معادله دینامیک پاندول معکوس خطی با ارتفاع ثابت است که به عنوان مدل حرکت ربات دوپا در تحلیل پایداری و طراحی مسیر ربات بسیار متداول است [10].

### فاز دو تکیه‌گاهی

فرض می‌شود فـاز دو تکیه‌گاهی بـه صـورت لحظـه‌ای رخ می‌دهـد به‌طوریکه همزمان با گذاشته شدن پای متحرک بر روی زمـین، پـای دیگر از زمین برداشته می‌شود. همچنین از ضربه کف‌پای متحـرک در برخورد با زمین صرف‌نظر می‌شود. از این رو در فاز دو تکیه‌گاهی بـا تغییر لحظه‌ای پای تکیه‌گاهی، مرکز فشار از یـک پـا بـه پـای دیگـر منتقل و به اندازه یک طول گام جابه‌جا می‌شود. بنابراین شرایط اولیـه گام جدید نسبت به شرایط نهایی گام کنونی عبارت خواهد بود از:

$$
\begin{cases}
x_{i+1}(t=0) = x_i(t=T_i) - L_i \\
\dot{x}_{i+1}(t=0) = \dot{x}_i(t=T_i)
\end{cases}
\quad (3)
$$

از طرفی جواب معادله (2) عبارت است از:

$$
\begin{cases}
x_i(t) = (e^{\omega t} + e^{-\omega t}) x_{i,0} + (\frac{e^{\omega t} - e^{-\omega t}}{\omega}) \dot{x}_{i,0} \\
\dot{x}_i(t) = \omega(e^{\omega t} - e^{-\omega t}) x_{i,0} + (e^{\omega t} + e^{-\omega t}) \dot{x}_{i,0}
\end{cases}
\quad (4)
$$

با جایگذاری حل فوق در معادلات (3)، معادله تفاضلی راه‌رفتن بـین شرایط اولیه گام‌ها برای مدل ساده شده حرکت، به صورت زیر بدست می‌آید:

$$\mathbf{x}_{i+1,0} = A\mathbf{x}_{i,0} + BL_i \quad (5)$$

که در آن:

$$
A = \frac{1}{2}\begin{bmatrix} e^{\omega T_i} + e^{-\omega T_i} & \dfrac{e^{\omega T_i} - e^{-\omega T_i}}{\omega} \\ \omega(e^{\omega T_i} - e^{-\omega T_i}) & e^{\omega T_i} + e^{-\omega T_i} \end{bmatrix}, B = \begin{bmatrix} -1 \\ 0 \end{bmatrix} \quad (6)
$$

---
[6] Swing leg protraction

## چرخه‌های حرکتی ساده

یک چرخه حرکتی ساده، دنباله‌ای از گام‌های کاملاً یکسان است. ازاین‌رو دنباله‌ای از گام‌ها با طول گام $L_c$ و زمان $T_c$ (در غیاب اغتشاش) یک چرخه حرکتی است اگر شرایط اولیه هرگام مشابه گام قبل باشد ($\mathbf{x}_{i+1,0} = \mathbf{x}_{i,0}$). باتوجه به معادله (۵) شرایط اولیه $\mathbf{x}_c$ یک چرخه حرکتی با طول گام $L_c$ و زمان $T_c$ به صورت زیر است.

$$\mathbf{x}_c = \frac{L_c}{2}\begin{bmatrix} -1 \\ \omega\dfrac{e^{\omega T_c}+1}{e^{\omega T_c}-1} \end{bmatrix} \tag{۷}$$

بنابراین، درصورتی که شرایط اولیه گام اول دقیقاً برابر (۷) اختیار شود، در غیاب اغتشاش دنباله‌ای از گام‌های یکسان به وجود می‌آید. برای بررسی پایداری حرکت در مقابل اغتشاش کافی است معادله خطای شرایط اولیه مورد بررسی قرار گیرد. معادله خطای شرایط اولیه با طول و زمان گام چرخه حرکتی با توجه به معادله گام به گام (۵) به شکل زیر بدست می‌آید.

$$\mathbf{e}_{i+1} = A_{T_i=T_c}\mathbf{e}_i \tag{۸}$$

مقادیر ویژه ماتریس $A_{T_i=T_c}$ عبارتند از $e^{\omega T_c}, e^{-\omega T_c}$ که یکی از مقادیر ویژه بزرگتر از یک است. بنابراین نقطه تعادل معادله (۸) ناپایدار است که نتیجه می‌شود چرخه‌های حرکتی ناپایدار بوده و اندک خطا در شرایط اولیه گام نسبت به شرایط اولیه چرخه حرکتی موجب افزایش خطا و دور شدن از چرخه حرکتی مطلوب می‌شود.

## پایدارسازی چرخه‌های حرکتی

انسان برای حفظ پایداری خود در مقابل ضربه و اغتشاش خارجی الگوی گام برداشتن (طول و زمان گام) خود را برای چند گام تغییر داده و پس از دفع اغتشاش به چرخه حرکتی پایدار خود برمی‌گردد. بنابراین تغییر طول و زمان گام‌ها می‌تواند راهکاری برای دفع اغتشاش و حفظ پایداری باشد. معادله گام به گام (۵) نیز نشان می‌دهد با تغییر طول و زمان هر گام می‌توان شرایط اولیه گام بعد را کنترل کرد. دراین صورت با طراحی کنترل‌کننده مناسب می‌توان شرایط اولیه را به شرایط اولیه چرخه مطلوب همگرا کرده و حرکت را حول آن پایدار کرد. در این مقاله صرفاً راهبرد تغییر طول گام مدنظر است و زمان گام‌ها ثابت و برابر زمان گام چرخه مطلوب درنظر گرفته می‌شود.

معادله خطای شرایط اولیه با ورودی تغییر طول گام نسبت به طول گام چرخه ($u_i = L_i - L_c$) به صورت زیر است.

$$\mathbf{e}_{i+1} = A_{T_i=T_c}\mathbf{e}_i + Bu_i \tag{۹}$$

### کنترل کننده تغییر طول گام

قانون کنترلی زیر درنظر گرفته می‌شود.

$$u_i = -K^T\mathbf{e}_i \tag{۱۰}$$

با استفاده از روش تثبیت قطب، ماتریس ضرایب کنترلی $K$ به صورت زیر محاسبه می‌شود.

$$\begin{cases} k_1 = \lambda_1 + \lambda_2 - 2A_{11} \\ k_2 = \dfrac{k_1 A_{11} - \lambda_1\lambda_2 + 1}{A_{21}} \end{cases} \tag{۱۱}$$

که در آن $\lambda_1$ و $\lambda_2$ قطب‌های سیستم حلقه بسته هستند. اگرچه قانون کنترلی (۱۰) پایداری سیستم حلقه بسته را تضمین می‌کند اما در عمل به خاطر محدودیت طول گام‌ها ممکن است ورودی محاسبه شده قابل اعمال نباشد. برای حل این مسئله می‌توان برای هرگام بسته به شرایط، قطب‌های جدید و درنتیجه ماتریس ضرایب جدید درنظر گرفته و برپایه یک الگوریتم جستجو بهترین آن‌ها را برای همگرایی در میان جواب‌های سازگار با محدودیت طول گام‌ها انتخاب کرد. در این مقاله به این موضوع پرداخته نمی‌شود.

### بهینه‌سازی ماتریس ضرایب کنترلی

در این بخش برای پرهیز از تولید ورودی کنترلی خارج از محدوده مجاز طول گام، ماتریس ضرایب کنترلی $K$ براساس بهینه‌سازی تابع هدف زیر بدست می‌آید.

$$J = \sum_{i=1}^{\infty}(\mathbf{e}_i^T Q\mathbf{e}_i + Ru_i^2) \tag{۱۲}$$

که در آن $Q$ یک ماتریس نیمه مثبت معین و $R$ یک عدد مثبت است. ماتریس ضرایب عبارت است از:

$$K = (R + B^T PB)^{-1}B^T PA \tag{۱۳}$$

که $P$ پاسخ معادله زیر است.

$$A^T PA - A^T PB(R + B^T PB)^{-1}B^T PA - P + Q = 0 \tag{۱۴}$$

با استفاده از تئوری تنظیم‌کننده‌های خطی مربعی، ثابت می‌شود که قانون کنترلی (۱۰) با ماتریس ضرایب (۱۳) سیستم (۹) را پایدار می‌کند. دراین حالت اگر ضریب $R$ به اندازه کافی بزرگ انتخاب شود، ورودی‌های کنترلی نزدیک به طول گام مطلوب انتخاب شده و در نتیجه در محدوده مجاز خواهند بود.

## شبیه‌سازی

به منظور بررسی عملکرد کنترل‌کننده‌های پیشنهادی تغییر طول گام، مدل ساده شده حرکت با پارامترهای جدول ۱ شبیه‌سازی شده است. سناریوی شبیه‌سازی بدین صورت است که حرکت با شرایط اولیه چرخه مطلوب شروع شده و در میانه گام سوم ضربه‌ای افقی به مرکز جرم ربات وارد می‌شود. در نتیجه ضربه وارد شده، شرایط اولیه گام بعد نسبت به شرایط اولیه چرخه خطا می‌یابد. براساس خطای ایجاد شده، کنترل‌کننده طول گام، طول گام‌های بعد از ضربه را تغییر داده تا شرایط اولیه به شرایط اولیه چرخه مطلوب همگرا شود. برای اختصار تنها نمودارهای کنترل‌کننده بهینه آورده شده‌اند. شکل ۲ نشان می‌دهد، پس از سه گام حرکت به چرخه حرکتی بازگشته و الگوی حرکتی قبل از ضربه تکرار می‌شود. مقابله گام به گام با اغتشاش و بازگشت به چرخه حرکتی با کنترل شرایط اولیه گام‌ها

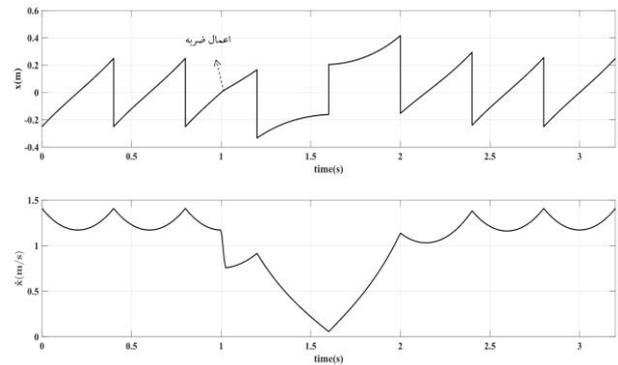

شکل ۲: نمودارهای پیوسته حرکت برای مولفه‌های مرکز جرم

(نقاط سیاه) در شکل ۳ مشخص است. خطوط افقی نشان‌دهنده فاز دو تکیه‌گاهی و آغاز گام جدید است. نتایج شبیه‌سازی نشان می‌دهد عملکرد کنترل‌کننده طراحی شده براساس راهبـرد تغییـر طـول گـام برای دفع اغتشاش و پایدارسازی حرکت موثر است. البته همانطورکـه در شکل ۴ دیده می‌شود با انتخاب ضریب $R$ بزرگتر در کنترل‌کننده بهینه، محدودیت طول گام‌ها به شکل بهتری رعایت می‌شود، هرچنـد سرعت همگرایی کمتر می‌شود.

جدول ۱: پارامترهای شبیه‌سازی

| پارامترهای راه‌رونده | |
|---|---|
| ارتفاع متوسط مرکز جرم | $h = 1\,m$ |
| شتاب جاذبه زمین | $g = 9.8\,m.s^{-2}$ |
| جرم کل | $m = 50\,kg$ |
| حداکثر اندازه طول گام | $L_{max} = 0.75\,m$ |
| پارامترهای چرخه حرکتی مطلوب | |
| مولفه موقعیت | $x_c = -0.25\,m$ |
| مولفه سرعت | $\dot{x}_c = 1.4\,m.s^{-1}$ |
| طول گام | $L_c = 0.5\,m$ |
| زمان گام | $T_c = 0.4\,s$ |
| پارامترهای ضربه اعمالی | |
| شدت نیرو | $F = -20\,N$ |
| مدت زمان اعمال نیرو | $\Delta T = 0.02\,s$ |

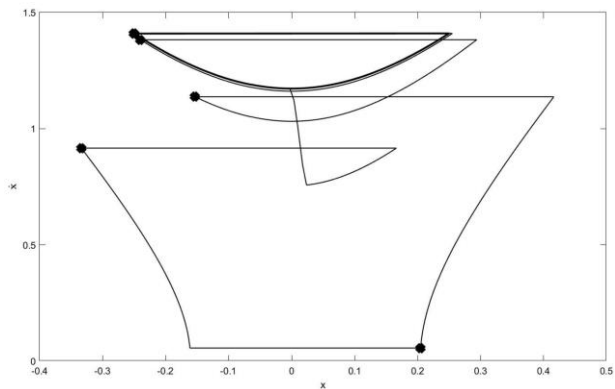

شکل ۳: صفحه فاز حرکت، بخش ضخیم نمودار چرخه حرکتی است.

حتی در مقابل اغتشاش شدید ضربه موثر است. این روش برای شـروع حرکت و رسیدن به یک چرخه مطلوب حرکتـی و همچنـین گـذار از یک چرخه حرکتی به دیگری (مـثلاً بـرای تغییـر سـرعت) هـم کاربرد دارد.

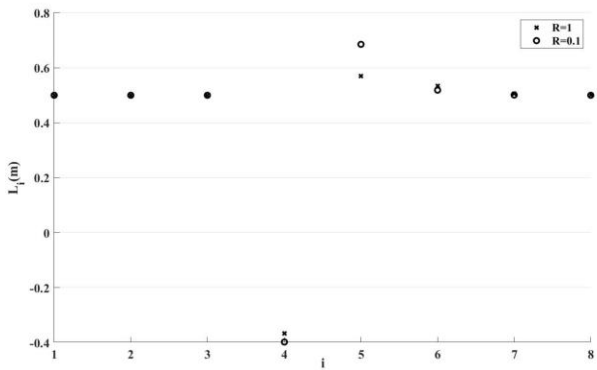

شکل ۴: طول هرگام برای دو مقدار ضریب $R$

## نتیجه‌گیری

در این مقاله پایدارسازی حرکت ربات دوپا با کنترل انـدازه‌حرکت‌های کلی ربات بررسی شد. در حقیقت نشان داده شد که براسـاس مفهـوم چرخه‌های حدی می‌توان معادله گام به گام حرکت را بـین شـرایط اولیه گام‌ها بدست آورده و برپایه آن با کنترل گسسته شـرایط اولیـه گام‌ها، حرکت را حول یک چرخه حدی پایدار کرد. برای این منظور، راهبرد تغییر طول گام بررسی و یک کنترل‌کننده خطی گسسته بـرای کنترل طول گام طراحی شد. در ادامـه بـرای تضـمین محـدودیت فیزیکی طول گام، ماتریس ضرایب کنترل‌کننده بهینه‌سازی شد. نتایج شبیه‌سازی نشان داد این روش در دفع اغتشاش‌ها و حفـظ پایـداری

## فهرست علائم

| | |
|---|---|
| $\mathbf{e}_i$ | بردار خطای شرایط اولیه گام $i$ نسبت به شرایط اولیه چرخه حدی |
| $f_x$ | مولفه افقی عکس‌العمل سطح |
| $f_y$ | مولفه عمودی عکس‌العمل سطح |
| $g$ | شتاب جاذبه زمین |
| $h$ | ارتفاع متوسط مرکز جرم از زمین |
| $H$ | اندازه‌حرکت زاویه‌ای کل حـول مرکـز جرم |
| $L_i$ | طـول گـام $i$ کـه عبـارت است از جابجایی مرکـز فشـار بـا تغییـر پـای تکیه‌گاهی |
| $m$ | جرم کل |
| $T_i$ | زمان گام $i$ |
| $x$ | موقعیت افقی مرکـز جـرم نسـبت بـه مرکز فشار ($x = x_{COM} - x_{COP}$) |
| $x_{COM}$ | موقعیت افقی مرکز جرم |
| $x_{COP}$ | موقعیت افقی مرکز فشار |
| $x_{i,0}$ | $x$ در ابتدای گام $i$ |
| $x_{i,T_i}$ | $x$ در انتهای گام $i$ |
| $\mathbf{x}_i$ | بردار شرایط اولیه گام $i$ |
| $y_{COM}$ | موقعیت عمودی مرکز جرم |

**علائم یونانی**

| | |
|---|---|
| $\omega$ | $\sqrt{g/h}$ |

**زیرنویس**

| | |
|---|---|
| $i$ | شماره گام |